\documentclass{article}

\usepackage{PRIMEarxiv}

\usepackage[utf8]{inputenc} 
\usepackage[T1]{fontenc}    
\usepackage{hyperref}       
\usepackage{url}            
\usepackage{booktabs}       
\usepackage{amsfonts}       
\usepackage{nicefrac}       
\usepackage{microtype}      
\usepackage{lipsum}
\usepackage{fancyhdr}       
\usepackage{graphicx}       
\graphicspath{{media/}}     

\usepackage{tabularx}

\newcommand{\lightweight}{TELL\space}

\title{Entity Linking in Tabular Data Needs the Right Attention
}

\author{%
  Miltiadis Marios Katsakioris \\
  School of Mathematical and Computer Sciences\\
  Heriot-Watt University\\
  Edinburgh, UK \\
  \texttt{mmk11@hw.ac.uk} \\
   \And
   Yiwei Zhou \\
   \texttt{yiwei1.zhou@gmail.com} \\
   \AND
   Daniele Masato \\
   Amazon Alexa \\
   Cambridge, UK \\
   \texttt{masatod@amazon.co.uk} \\
}

\begin{document}
\maketitle

\begin{abstract}
Understanding the semantic meaning of tabular data requires Entity Linking (EL), in order to associate each cell value to a real-world entity in a Knowledge Base (KB). In this  work, we focus on end-to-end  solutions for EL on tabular data that do not rely on fact lookup in the target KB. Tabular data  contains heterogeneous and sparse context, including column  headers, cell values and  table captions. 
We experiment with various models to generate a vector representation for each cell value to be linked. Our results show that it is critical to apply an attention mechanism as well as an attention mask, so that the model can only attend to the most relevant context and avoid information dilution. The most relevant context includes: same-row cells, same-column cells, headers and caption. Computational complexity, however, grows quadratically with the size  of tabular data for such a complex model. We achieve constant memory usage by introducing a Tabular Entity Linking Lite model (\lightweight) that generates vector representation for a cell based only on its value, the table headers and the table caption. \lightweight\ achieves 80.8\% accuracy on Wikipedia tables, which is only 0.1\% lower than the state-of-the-art model with quadratic memory usage.
\end{abstract}

\section{Introduction}

Tabular data, such as web tables and databases, provides invaluable information about the world.
According to \cite{cafarella2018ten}, by 2008, there are 14.1 billion HTML tables from Google’s general-purpose web crawl, and 154M of them are high quality relational data.
There has been various efforts to leverage rich factual information contained in tabular data for Knowledge Base Augmentation \cite{ritze2016profiling,kruit2019extracting}, Question Answering \cite{chen2020open}, etc.
These applications require to automatically interpret and understand the semantic meaning of tabular data at scale.
Entity Linking in tabular data, which target at linking a cell value (``\texttt{Titanic}'') in tabular data with its corresponding real-world entity reference in a knowledge base (Q44578 in Wikidata\footnote{\url{https://www.wikidata.org/}}), is an important step for semantic table interpretation.

Comparing with Entity Linking in unstructured text \cite{logeswaran2019zero,martins2019joint}, Entity Linking in tabular data needs to tackle some additional challenges.
First, each cell value is an entity mention to be linked, the environment it appears is not a complete sentence, but other entity mentions in short texts, real numbers, or dates.
Second, besides content information in the form of rows and columns, tabular data is usually associated with metadata information, such as HTML page titles, table captions and column headers. For each cell in the table, this metadata information can contain both signals and noises. 
Third, sizes of tables can vary significantly. Once the table schema is determined, the size of a table can grow infinitely. 

In this work, we focus on end-to-end Entity Linking solutions in tabular data, the main contributions are as follows:

\begin{itemize}
    \item Besides entity disambiguation, we also consider the scenarios that a cell value is not an entity mention and the correct entity is not included in the list of candidate entities, so that the solution is robust to errors introduced during candidate entities retrieval.
    \item We reduce the required prior knowledge for existing entities in the targeted KB to entity names and entity descriptions, which increases the solution's generalisation capability to unseen entities.
    \item We experiment with various models to generate a vector representation for each cell value to be linked, and verified the importance of applying an attention mechanism and an attention mask to regulate the interactions between cell values and table metadata.
    \item We propose a simple yet effective model, Tabular Entity Linking Lite (TELL), which generates cells' vector representations only based on cell values and metadata. TELL reduces the computation complexity of tabular structure aware models from quadratic to linear while sacrificing the accuracy by 0.1\%, from 80.9\% to 80.8\%.
\end{itemize}

\section{Related Work}
Most former works \cite{ritze2016profiling,kruit2019extracting} on Entity Linking in tabular data requires fact lookup during training and inference time. 
These solutions can only link cell values with entities that are well populated with facts in the targeted KB, and their entity linking capability is restricted to only cell values in the subject column of a table. There has been some learning based Entity Linking approaches for tabular data. However, they are either based on an assumption that the correct entity is included in the list of retrieved candidate entities, and only focus on the disambiguation part of the problem \cite{deng2020turl,luo2018cross}; or they are dependent on some additional knowledge about the entities in the KB, such as prior probabilities of entity mentions \cite{53f99c216b614a02a2cf31bff30895d0}, entity embeddings \cite{luo2018cross}, or entity types \cite{deng2020turl}, which limit their generalisation capability to unseen entities during training.

\begin{figure}[]
\centering
\includegraphics[height=10cm]{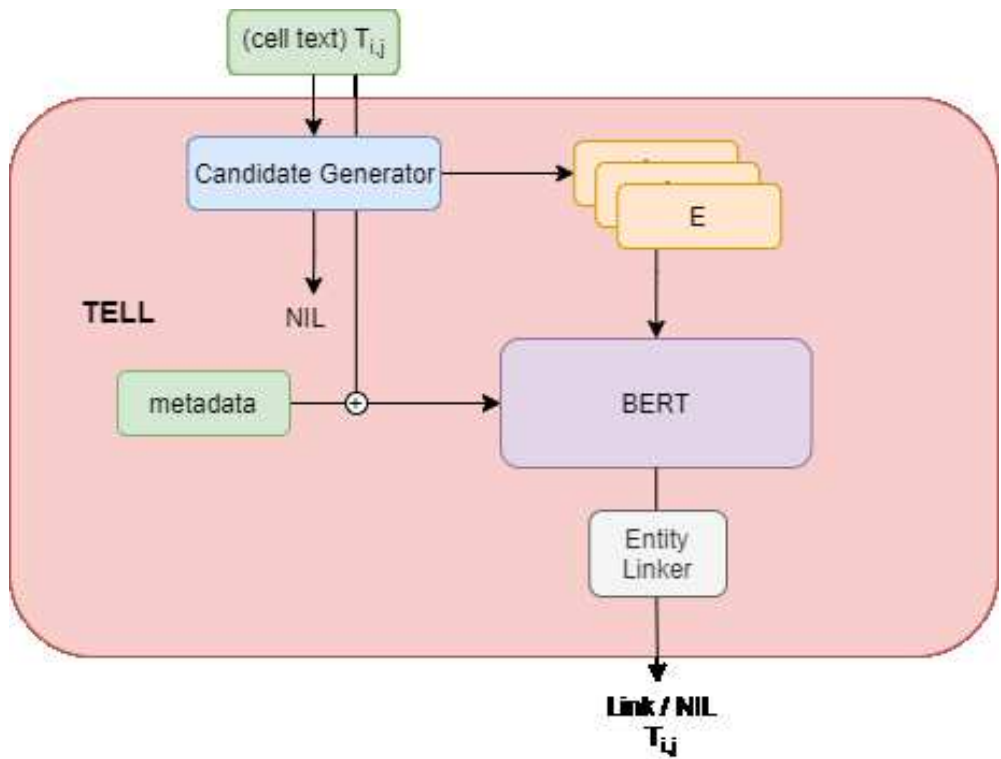}
\caption{Overview of our EL pipeline. First, all entity mentions $T_{i,j}$ are passed through the candidate generator. The resulting list of candidates $E$ and the $T_{i,j}$ with the corresponding metadata are encoded by BERT and the entity linker, ranks and makes the final decision for each $T_{i,j}$, whether it has a link from $E$ or not.}
\label{fig:pipeline}
\end{figure}

\section{Entity Linking in Tabular Data}
Without loss of generality, we assume that each table $T$ consists of $M$ data rows and $N$ data columns, and $T_{i,j}$ represents the cell value in the $i^{th}$ row and $j^{th}$ column. Each table can be associated with additional metadata, which includes: (1) Table caption $C$, a short text description summarizing the content of the table; (2) Page title $P$, the title of the web page the table was in; (3) Table headers $H = [h_0, \dots h_{N-1}]$, one for each column to define the table schema. 

Same as \cite{luo2018cross,deng2020turl,53f99c216b614a02a2cf31bff30895d0}, we consider each table cell $T_{i,j}$ is a \textit{potential} mention for real world entities in the target KB.
Specifically, we define the Entity Linking in tabular data problem as follows:
\newline
\newline
\newline
\textit{Definition 1.} Given a table $T$ and a target KB, link cells $T_{i,j}$ in $T$ with their corresponding real-world entity references in the target KB if possible, while automatically detect and ignore the ones that cannot be linked. 

To increase the entity linking solution's generalisation capability to unseen entities \cite{logeswaran2019zero} and any target KB,  we only assume the existence of names and descriptions for entities in the target KB, which is minimal comparing with former works.

Figure~\ref{fig:pipeline} 
presents an overview of our EL pipeline which consists of two stages: (1) Candidate Generation, which generates candidate entities $E$ for all cells in each table $T$, and (2) Entity Disambiguation, which ranks and selects the best $e \in E$ for each $T_{i,j}$ if there is any (refer to as NIL otherwise).

\subsection{Candidate Generation}
\label{sec:CR}
Most existing work~\cite{53f99c216b614a02a2cf31bff30895d0, 10.1007/978-3-319-68288-4_16, 10.1145/2797115.2797118, deng2020turl} directly use exact string matching-based lookup services provided by the target KB to generate candidate entities for each cell, which cannot tackle the variations of entity mentions in web tables. A more sophisticated method for this stage, we can increase the probability of including the right entity in the candidate entities ($P_E$) while introducing minimal noises. Following \cite{ganea2017deep}, we used an gazetteer constructed based on Wikidata entity names and alias, as well as Wikipedia article titles, hyperlinks and redirects. Additionally, we applied BM25 to measure the similarity between a cell value and a KB entity rather than exact string matching.
According to our analysis, the combination of gazetteer and BM25-based similarity search has increased $P_E$ from
77\% to 88\%.

\subsection{Entity Disambiguation}

We use a shared BERT~\cite{devlin-etal-2019-bert} encoder to encode all the textual inputs, which include: cell value $T_{i,j}$, metadata $(C, P, H)$, entity name $e_{name}$ and entity description $e_{desc}$.

For a KB candidate entity, its vector representation $e$ is achieved by adding its averaged name token embedding and its averaged description token embedding:
\begin{equation}
    e = mean(BERT(e_{name})) + mean(BERT(e_{desc}))
\end{equation}



When encoding the cells of a table we have to take into account how to model the tabular structure, if and how to use metadata and the role of attention for the optimal representation. We compare different ways of encoding in a top-down approach.
At the top, in terms of complexity and information load is $MaskAttEnc$, an encoder based on TURL~\cite{deng2020turl}, a state-of-the-art framework for relational table understanding that consists of a structure-aware Transformer encoder to model the row-column structure of the table and capture the textual information and relational knowledge of each cell. During the self-attention calculation, a ``hard-coded'' attention mask limits the aggregation of information from one entity cell to other structurally related entity cells, such as cells in the same row/column. In our ablation study, when we are not using the hard-coded attention mask, we refer to the model as $AllAttEnc$.

We remove the hard-coded attention mask and we treat each cell value as a separate entity. Instead of encoding the metadata separately using extra attention heads and fusing them with the encoded cells, we concatenate them to each cell. We refer to the resulting module as \lightweight. $MaskAttEnc$ makes predictions on the table level whereas \lightweight~can make predictions on the individual cell level, allowing for extra flexibility. 

We simplify the entity mention representation of \lightweight further by removing all BERT attention heads. We refer to these models as $SingleEnc$, with the cells being treated separately as in TELL. However, instead of attention, we encode the sentence embeddings either by an LSTM, $SingleLSTMEnc$ or by averaging the embeddings and passing them through a linear feedforward layer, $SingleLinearEnc$.



After getting the vector representation for a cell, we calculate the matching score between $T_{i,j}$ and $e$ by,
\begin{equation}
    P(e) = \frac{exp(T_{i,j}\cdot e)}{\sum_{e^{'}\epsilon E}exp(T_{i,j}\cdot e^{'})}
\end{equation}
and we select the candidate with the highest probability.

\section{Dataset}
For our experiments we pre-processed and generated our own data splits of the WikiTables corpus~\cite{53f99c216b614a02a2cf31bff30895d0}.
WikiTable corpus contains 1.65M tables extracted from Wikipedia pages, and most of the tables contain hyperlinks between cell values and Wikipedia entity articles labelled by Wikipedia contributors. 

We cleaned the table cells by lower-casing their content, removing HTML tags and removing special characters. For cells containing multiple hyperlinks, we retained the first link only. 
In order to fit training batches (25 batches) in memory, we discarded tables with more than 500 cells (entity mentions). 
We also discarded tables with no linked entities, and tables 
with either no candidates or more than 1800 candidates overall.
For computational efficiency we remove duplicate cell values from the tables.
In addition, to remove noises in Wikipedia hyperlinks, we compute the difference in lengths between each cell value and its corresponding linked entity's name. As long as the difference is bigger than 10, we will ignore the hyperlink because it is very unlikely to happen.

In order to map hyperlinked Wikipedia entities in WikiTables with Wikidata entities, we use Wikimapper\footnote{See \texttt{https://github.com/jcklie/wikimapper}.}.


\begin{table}
\centering
\begin{tabular}{lllll}
\hline Splits & \textbf{\# Tables} & \textbf{\# NIL} & \textbf{\# Entities} \\ \hline
Train & 554,239 & 16.6M & 26.6M\\
Validation & 4,738 & 142K & 317K \\
Test & 4,660 & 139K & 312K\\
\hline
\end{tabular}
\caption{\label{data-stats} Dataset statistics. In the `NIL' column we show an estimate of the entity mentions that have no link and in the last column the total entity mentions $T_{i,j}$ from all tables. }
\end{table}


Statistics of our data splits are summarized in Table~\ref{data-stats}. The table shows that the mean percentage of NIL entities in all splits is around 50\%. The average number of NIL entities is consistent for both the dev and test splits.



\section{Experimental Results}
We evaluated our solution end-to-end using accuracy and F1 score. 
We first established a strong baseline, by encoding each cell, only with the cell value using BERT.
This baseline, which does not use any extra context, achieves an accuracy of 77.5\%.
This baseline is simply learning mappings from cell value embeddings to entity embeddings, we expect any approach that uses additional context to outperform the baseline. 

Table~\ref{results} shows increasing levels of ablation. It starts with $MaskAttEnc$, a model using both the whole table content and the metadata, then transitions to the aforementioned baseline model, a model using only separated cell values.
The proposed solution, \lightweight, uses separated cell values as well as metadata,
achieves  80.8\% accuracy and 79.3\% F1 score, and it only requires a linear number of parameters. 
The comparison between \lightweight and the baseline shows the importance of the metadata and the attention that happens in each cell between its text and the metadata.

On the other hand, $MaskAttEnc$ attends to the whole table and achieves 80.9\% accuracy with the help of the attention mask.
This mask regulates that each cell only attend to structurally related cells from the entire table and the metadata.
We see the importance of the attention mask when we compare it with $AllAttEnc + meta$ with a drop in accuracy of 2.3\%. 
Without the attention mask as a regularization method, the model fails to automatically learn the most relevant information for each cell.

A marginal improvement (0.1\%) in accuracy compared to \lightweight, for the sacrifice of scalability since computation complexity of the $MaskAttEnc$ solution
grow exponentially $O(n^2)$. 
This is because $MaskAttEnc$ needs to model the relationship between any pair of cell values, rather than treating them separately as TELL.
This also makes $MaskAttEnc$ able to only work on small tables. 
For~\lightweight and its ablations, 
table size is not an issue,
as long as the metadata are being passed on together with each cell. 
Aside from the type of attention and the structured relatedness, we see that the metadata are crucial and always contribute to the performance.




For tabular data is critical to apply the right attention
in order to attend to the right cells when generating the cell representations. Otherwise, information dilution from irrelevant cells will greatly impact the performance and it might be better to simply pass each cell individually.


\begin{table}
\centering
\begin{tabular}{lllll}
\hline \textbf{Method}  & \textbf{Acc} & \textbf{F1} & \textbf{Comp.} \\ \hline
\small MaskAttEnc + meta (TURL) & \textbf{80.9} & \textbf{80.3} & $O(n^2)$\\ 
\small SingleAttEnc + meta (\lightweight) & \textit{80.8} & \textit{79.3} & $O(n)$\\
\small AllAttEnc + meta & 78.6 & 78.1 & $O(n^2)$\\
\small SingleLinearEnc + meta & 77.9 & 76.1 & $O(n)$\\
\small SingleAttEnc (baseline) & 77.5 & 76.3 & $O(n)$\\ 
\small AllAttEnc & 76.8 & 76.0 & $O(n^2)$\\
\small SingleLSTMEnc + meta & 69.7 & 69.1 & $O(n)$\\ \hline
\end{tabular}
\caption{\label{results} Evaluation of all variants on test set. In the Big $O$ notation, $n$ symbolises the number of input cells.}

\end{table}

\section{Conclusion}

The state-of-the-art framework for relational table understanding (TURL) performs EL for a given cell using a Deep Neural Network with attention that aggregates information from the whole table, including the content and location of surrounding cells, and additional table metadata such as title, caption and headers. TURL achieves 80.9\% accuracy on Wikipedia tables, but its computation complexity grows quadratically with the number of cells in a table.

In this paper we presented a lightweight approach for EL on tabular data that can achieve almost state-of-the-art accuracy with linear computation complexity. 
We consider both challenges of candidate retrieval and entity disambiguation, whilst trying to find a balance between the two. We showed that metadata are crucial. In order to avoid a noisy cell representation it is important to apply the right attention to avoid information dilution.
In future work, we will focus on improving the candidate entity retrieval mechanism, by applying some embedding based approach.
Another option is to develop solutions to transform tabular data into unstructured text, in order to leverage EL solutions for unstructured text to tackle the problem.






\bibliographystyle{unsrt}  
\bibliography{anthology}

\end{document}